\documentclass[final,1p,times]{elsarticle}

\usepackage{graphicx}
\usepackage{amssymb}
 \usepackage{amsthm}
\usepackage{amsmath}
\usepackage{lineno}
\usepackage{multirow}

\biboptions{numbers,sort}

\journal{arXiv}

\begin{document}

\begin{frontmatter}


\title{Peristaltic locomotion without digital controllers:
Exploiting the origami multi-stability to coordinate robotic motions}



\author{Priyanka Bhovad\corref{cor1}}
\cortext[cor1]{Corresponding Author: pbhovad@g.clemson.edu}
\author{Joshua Kaufmann}
\author{and Suyi Li}

\address{Department of Mechanical Engineering, Clemson University, Clemson, SC, USA}

\begin{abstract}
This study proposes and examines a novel approach to generate peristaltic-like locomotion in a segmented origami robot.  Specifically, we demonstrate the use of multi-stability embedded in origami skeleton to eliminate the need for multiple actuators or digital controllers to coordinate the complex robotic movements in peristaltic crawling.  The crawling robot in this study consists of two serially connected bistable origami segments, each featuring a generalized Kresling design and a foldable anchoring mechanism. Mechanics analysis and experimental testing of this dual-segment module reveal a deterministic deformation sequence or actuation cycle, which is then used to generate the different phases in a peristaltic-like locomotion gait.  Instead of individually controlling the segment deformation like in earthworm and other crawling robots, we only control the total length of this robot.  Therefore, this approach can significantly reduce the total number of actuators needed for locomotion and simplify the control requirements. Moreover, the richness in Kresling origami design offers us substantial freedom to tailor the locomotion performance.  Results of this study will contribute to a paradigm shift in how we can use the mechanics of multi-stability for robotic actuation and control.

\end{abstract}


\end{frontmatter}


\section{Introduction}
\begin{figure}[h!]
\centering\includegraphics[]{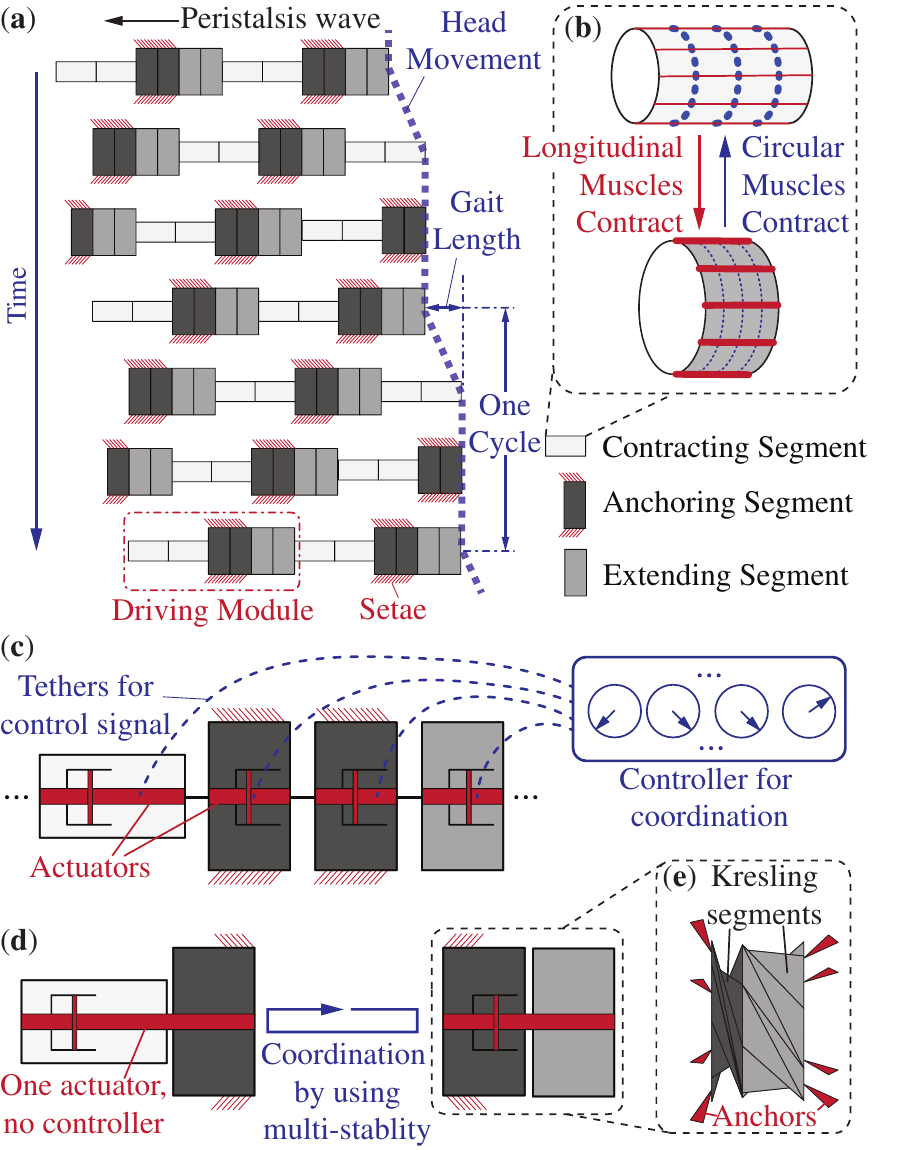}
\caption{The vision of using multi-stability to drastically simplify the mechatronic setup for generating peristaltic locomotion.  (a) Peristaltic locomotion cycle in an earthworm. The earthworm body moves forward while the peristaltic wave propagates backwards. For clarity, the earthworm body consists of six identical segments and two driving modules. (b) The muscular actuation scheme of an earthworm segment. The alternate contraction of longitudinal and circular muscles drives the segment deformation and anchoring actions. (c) The mechatronic setup of a traditional earthworm-inspired robot that requires many actuators and a complicated controller. (d) The proposed peristalsis locomotion mechanism that uses multi-stability to eliminate the need of multiple actuators and controllers. (e) A to scale schematic diagram of the dual-Kresling driving module and foldable anchors. }
 \label{fig:Bigpic}
\end{figure}

Limbless and metameric invertebrates like the earthworm use peristalsis to crawl over uneven surfaces, burrow through soil, and navigate in confined spaces with ease.  The body of an earthworm consists of many segments that are grouped into several ``driving modules''.  Each module includes three types of segments according to their states of deformation: ``contracting'', ``anchoring'', and ``extending'' \cite{Quillin1999} (Figure \ref{fig:Bigpic}(a)).   In a peristaltic locomotion cycle, the contracting segment expands in diameter and contracts in length by engaging its longitudinal muscles (Figure \ref{fig:Bigpic}(b)).  The extending segment deforms in the opposite way by engaging its circular muscles.  When a contracting segment reaches the fully-contracted shape, it becomes an anchoring segment, which can firmly attach itself to its surrounding by further deploying hair-like bristles (aka. \textit{setae}) on its surface.  By carefully {\it coordinating} the deformation of its segments, the earthworm can generate a retrograde peristaltic wave that propagates towards the tail end of its body, thus driving itself forward (Figure \ref{fig:Bigpic}(a)).  

The locomotion performance of a peristaltic gait can be easily tuned by changing the number of these three types of segments in a driving module \cite{Fang2015,Fang2015b}. The absence of complex external appendages like legs or wings makes the driving module design compact and light. As a result, peristaltic locomotion has been implemented in many worm-inspired crawling robots for field exploration and in-pipe inspection.  However, these robots typically require many actuators---such as pneumatic chambers \cite{Calderon2019, Kamata2018, 8594390}, shape memory alloy (SMA) springs \cite{Fang2017}, electric motors \cite{Fekrmandi2019}, or permanent magnets \cite{Saga2004}---to activate their segments {\it individually}.  Moreover, a complicated control architecture is also necessary to coordinate the individual segment deformation to achieve peristaltic locomotion (Figure \ref{fig:Bigpic}(c)).  This can lead to a cumbersome mechatronic setup that can significantly constrain the overall application potential, especially when these robots need to be completely soft and un-tethered \cite{Rich2018}.

To address this issue, we examine the use of {\it multi-stability} in Kresling origami to generate peristaltic-like locomotion without relying on multiple actuators or digital controllers.  A material or structure is multi-stable when it possesses more than one stable equilibria (or states).  It can remain at one of its stable states without any external aid, and switch between these states by external or internal actuation. Multi-stability is a catalyst for creating smart materials and structures that can provide energy harvesting \cite{Harne2013, Daqaq2014, Younesian2017}, vibration isolation \cite{Johnson2013, Johnson2014, Hu2015}, shape morphing \cite{Panesar2012, Sun2016}, and even non-reciprocal wave propagation \cite{Nadkarni2016, Raney2016}.

Regarding its applications in robotics, multi-stability also shows promise in amplifying the authority and speed of robotic actuation \cite{Fang2017, Yang2015, Kim2014}, or increasing the precision and repeatability of a micro-robotic end effector \cite{Chalvet2013}.  More importantly, recent studies reveal that multi-stability can be harnessed to drastically reduce or even eliminate the need for using digital controllers to generate locomotion \cite{Rafsanjani2019}.  For example, multi-stability was used to pre-program the flapping of robotic fins to achieve forward and backward swimming \cite{Chen2018a}. It was also used to induce coordinated motions for non-peristaltic crawling \cite{Suo2018, Treml2018} and gripping \cite{Preston2019} without using any digital controllers. Buckling---a characteristic behavior induced by multi-stability---was exploited to sequence robot leg motions for walking \cite{Gorissen2019}.

In this study, we show that by exploiting the multi-stability from origami, we can create peristaltic-like crawling locomotion with only one actuator and without any digital controllers (Figure \ref{fig:Bigpic}(d)).  The crawling robot in this case consists of a driving module of two serially connected {\it Kresling segments} and foldable anchors (Figure \ref{fig:Bigpic}(e)).  We designed the Kresling pattern according to the desired kinematics and bistability so that these segments can exhibit both longitudinal and radial deformation via folding.  When its total length is increased and decreased by a linear actuator, the dual-Kresling driving module can exhibit a deterministic deformation sequence (or ``actuation cycle'') between its different stable states.  We then designate different parts of this actuation cycle as the phases of a peristaltic-like locomotion gait.  By doing so, we can eliminate the need for using individual actuators for each segment or using digital controllers to coordinate these actuators. That is, the peristaltic locomotion is essentially ``coordinated'' by the mechanics of multi-stability in Kresling origami.  Therefore, this study will contribute to a paradigm shift in how we can use the mechanics of multi-stability for robotic actuation and control.

The following sections of this letter will (2) detail the design, analysis, and experimental characterization of the elementary Kresling origami; (3) elucidate the creation of a deformation sequence (or ``actuation cycle'') using origami multi-stability; (4) discuss the design and experimental validation of the peristaltic-like locomotion using this actuation cycle; and (5) conclude this study with summary and discussion. 

\section{Generalized Kresling Origami}
The center piece of the peristaltic crawling robot in this study is a driving module consisting of two serially connected Kresling origami segments.  Origami is an ancient art of paper folding wherein folding a 2D sheet along prescribed crease lines results in the creation of complex 3D shapes.  Over the past few decades,  it has become a framework for constructing deployable structures \cite{Zirbel2013}, mechanical metamaterials \cite{Li2018}, and reconfigurable robots \cite{Rus2018}.  Origami mechanisms are inherently lightweight, compact, and compliant.  Moreover,  they can exhibit unique mechanical properties---such as auxetics, programmable nonlinear stiffness, and multi-stability \cite{Filipov2015, Li2015, Li2015a, Fang2016, Schenk2011, Yasuda2015, Sadeghi2019}---due to the nonlinear kinematics of folding.  Here, we choose Kresling origami as the skeleton of our peristaltic crawling robot.  

\begin{figure}[h!]

\makebox[\textwidth][c]{\includegraphics{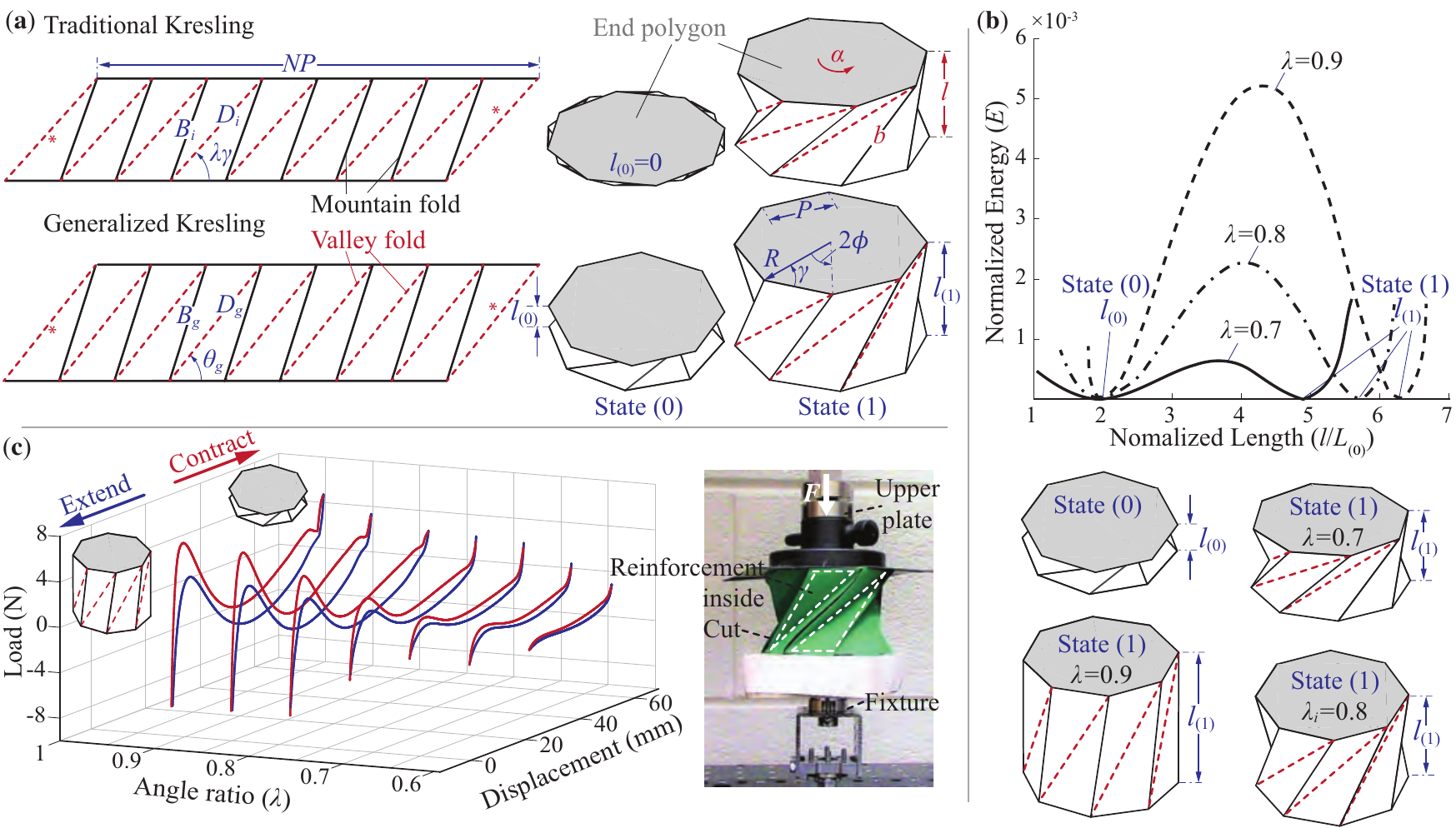}}

\caption{Design, analysis, and experimental characterization of the generalized Kresling origami.  a) Crease pattern and the folded segment of both traditional and generalized Kresling origami showing the important design parameters and variables related to folding.  The traditional Kresling always has a zero length at the fully-contracted state (0), while the generalized Kresling has a ``user-defined'' $l_{(0)}$.  b) The normalized strain energy versus length of three Kresling segment designs of different angle ratios but the same $l_{(0)}(=20$mm), $N(=8)$, and $P(=30$mm).  Increasing the angle ratio can increase the bistability strength and length of segment at the fully-extended state (1).   c) Experimentally measured force-displacement curves of Kresling segments with different angle ratios but the same $l_{(0)}(=10$mm), $N(=8)$, and $P(=30)$mm.  One can clearly see the correlation between angle ratio and bistability strength in terms of the maximum reaction force between stable states. The inserted picture on the right shows the experimental setup.}
 \label{fig:Kresling}
\end{figure}

The Kresling pattern consists of a linear array of mountain and valley folds defined by triangular facets (Figure \ref{fig:Kresling}(a)). By attaching the two ends of this array (marked by *), we obtain a twisted polygonal prism with a regular polygon at its top and bottom. These two end polygons remain rigid throughout the folding motion. Kresling origami was originally studied as a buckling mode in thin cylindrical shells subjected to torsion. Since then, it has been used extensively as a template for deployable structures or robotic skeletons \cite{bhovad2018using, guest1994folding, hunt2005twist, jianguo2015bistable, kresling2008natural, Nayakanti2017, pagano2017crawling}. Kresling origami suits this study well because it has the desired tubular cross-section, and more importantly, it is inherently bistable.  A Kresling segment can settle in a fully-extended or a fully-contracted stable state, and it exhibits a large deformation between these two states. This bistability originates from its non-rigid-foldable nature.  The triangular facets remain undeformed at the two stable states, but must deform while folding between these two states. Indeed, if these triangular facets were strictly rigid, the Kresling segment would not fold. For clarity, we refer the fully-contracted stable state as the state (0) and the fully-extended stable state as the state (1) hereafter.

The design of a {\it traditional} Kresling segment is fully defined by three independent parameters:  the number of sides of the base and top polygon $N$,  the side length of the polygon $P$, and an angle ratio $\lambda$, which is related to the angle between polygon side and valley crease in the triangular facets (Figure \ref{fig:Kresling}(a)).  The length of the valley and mountain creases are:
\begin{align}
   D_i&=2R\cos(\gamma-\lambda \gamma),\\
   B_i&=\sqrt {P^2+D_i^2-2 P D_i\cos(\lambda \gamma)},
\end{align}
where $\gamma \, (=\pi/2-\phi)$ is the angle between the diagonal and side of the end polygon, $R \, (=0.5 P/\sin\phi)$ is its circumscribed radius, and $\phi=\pi/N$. The traditional Kresling design, however, has a shortcoming: Its length at the fully-contracted stable state (0) is always zero.  This is impossible in practice due to the finite material thickness, more importantly, it significantly constrains the design space available for tailoring the kinematics of peristalsis crawling.  To address this issue, we created a ``generalized'' Kresling pattern by adding the fourth independent design variable: a non-zero segment length at stable state (0) (aka. $L_{(0)}$ in Figure \ref{fig:Kresling}(a)) \cite{bhovad2018using}. The triangular facets are ``stretched'' as a result and their geometry is adjusted accordingly:
\begin{align}
    D_g&=\sqrt{D_i^2+L_{(0)}^2},\\
    B_g&=\sqrt{B_i^2+L_{(0)}^2}.\\
    \theta_g&=\cos^{-1}\left(\frac{P^2+D_g^2-B_g^2}{2PD_g}\right)
\end{align}
Here, $\theta_g$ is the angle between polygon side and valley crease, the subscript ``{\it i}'' denotes the traditional Kresling and ``{\it g}'' denotes the generalized Kresling. By using this generalized design, we can freely assign {\it non-zero} lengths to the Kresling segment at both fully-contracted state (0) and fully-extended stable state (1). 

To characterize the bistability of generalized Kresling segments, we adopt the equivalent truss frame approach.  This approach uses pin-jointed truss elements to represent the mountain and valley creases and assumes that the valley creases do not change their length during folding \cite{jianguo2015bistable, guest1994folding}.  In this way, the triangular facet deformations induced by folding between the two stable states can be approximated as the stretching and compression of the truss elements along mountain creases.  More specifically, the mountain crease trusses are un-deformed at the two stable states, but they are compressed as we fold the Kresling segment between its two states. To describe the Kresling folding deformation, we use three variables: the relative rotation angle between the top and bottom end polygon during folding ($\alpha$), the overall length of the Kresling segment $l$, and the length of the truss element along mountain creases $b$. These three variables are applicable to both traditional and generalized Kresling, and they are inter-dependent. Especially, the values of $\alpha$ are the same between the traditional and generalized Kresling, so we can use it as the independent variable and obtain a closed form solution describing the folding kinematics:
\begin{align}
    l(\alpha)&=\sqrt{L_{(0)}^2+2R^2\left[\cos(\alpha+2\phi)-\cos(\alpha_{(0)}+2\phi)\right]},\label{eq:L}\\
b(\alpha)&=\sqrt{2R^2(1-\cos(\alpha))+l^2}.
\end{align}
Here, $\alpha_{(0)} (=2\lambda \gamma)$ is the angle between top and bottom polygon at the fully-contracted stable state (0), and $l=L_{0}$ at this state according to Equation (\ref{eq:L}).  Angle $\alpha$, corresponding to the fully-extended stable state (1), can be computed by setting the mountain crease length $b$ equal to its undeformed length $B_g$: 
\begin{equation}
    \alpha_{(1)}=\lbrace \min(\alpha)|b(\alpha)=B_g \rbrace.
\end{equation}
The equivalent strain ($\epsilon$) and strain energy ($U$) due to folding are 
\begin{equation}
    \epsilon=\frac{b}{B_g}-1\text{ and }U=\frac{1}{2}K\epsilon^2,
\end{equation}
where $K$ represents the constituent sheet material stiffness. For the purpose of this analysis, we normalize the strain energy $U$ by $K$, and define the non-dimensional strain energy as
\begin{equation}
E=\frac{1}{2}\epsilon^2. \label{eq:E}
\end{equation}
Figure \ref{fig:Kresling}(b) illustrates the normalized strain energy of three Kresling designs with the same $L_{(0)}$ but different angle ratios $\lambda$.  The two potential energy wells are evident in these analytical results.  Moreover, as the angle ratio increases, the effective strain $\epsilon$ increases, consequently increasing the bistability strength in terms of the height of energy barrier between stable states.  For a given $L_{(0)}$, the bistablity is weakest when $\lambda=0.5$ and strongest when $\lambda=1$. The Kresling segment is no-longer bistable if $\lambda<0.5$. 

We fabricated prototypes of the generalized Kresling segments using paper (Daler - Rowney Canford 150 gsm) and experimentally validated their bistability.  We first prepared the 2D drawing of Kresling pattern in SOLIDWORKS\texttrademark and cut them out of paper with perforated creases on a plotter cutter (Cricut Maker$^{\circledR}$).  We then manually folded the cut pattern into the Kresling segment and attached its top and bottom polygons to the universal test machine (ADMET eXpert 5601).  To accommodate the relative rotation of these end polygons, we designed a custom rotation fixture consisting of a dual ball-bearing hub (Figure \ref{fig:Kresling}(c)).  Certain adjustments to the Kresling segment fabrication were necessary to facilitate smooth folding.  First, we cut the mountain creases to alleviate any excessive stresses that can lead to tearing after a few loading cycles.  A similar approach is used in the  ``Flexigami'' \cite{Nayakanti2017}. Secondly, we added triangular reinforcements to the facets to increase their stiffness relative to the creases, strengthening overall bistability (Figure \ref{fig:Kresling}(c)). 

Figure \ref{fig:Kresling}(c) also illustrates the measured force-displacement curves of several Kresling segment prototypes.  The correlation between angle ratio and bistability strength is evident in that a segment with a higher angle ratio demands a larger actuation force to be switched between stable states.  Moreover, we observe a hysteresis loop between the extension and contraction cycles.  This hysteresis behavior is intrinsic to the system, and it probably originates from the delamination in the paper along crease lines and the contact between triangular facets.  Nonetheless, we can minimize this hysteresis by the aforementioned cutting and reinforcement techniques so that it will not significantly affect the generation of the actuation cycle as discussed below. 

\section{Actuation Cycle from the Multi-stable Driving Module}\label{Act_cycle}
\begin{figure}[h!]

\makebox[\textwidth][c]{\includegraphics{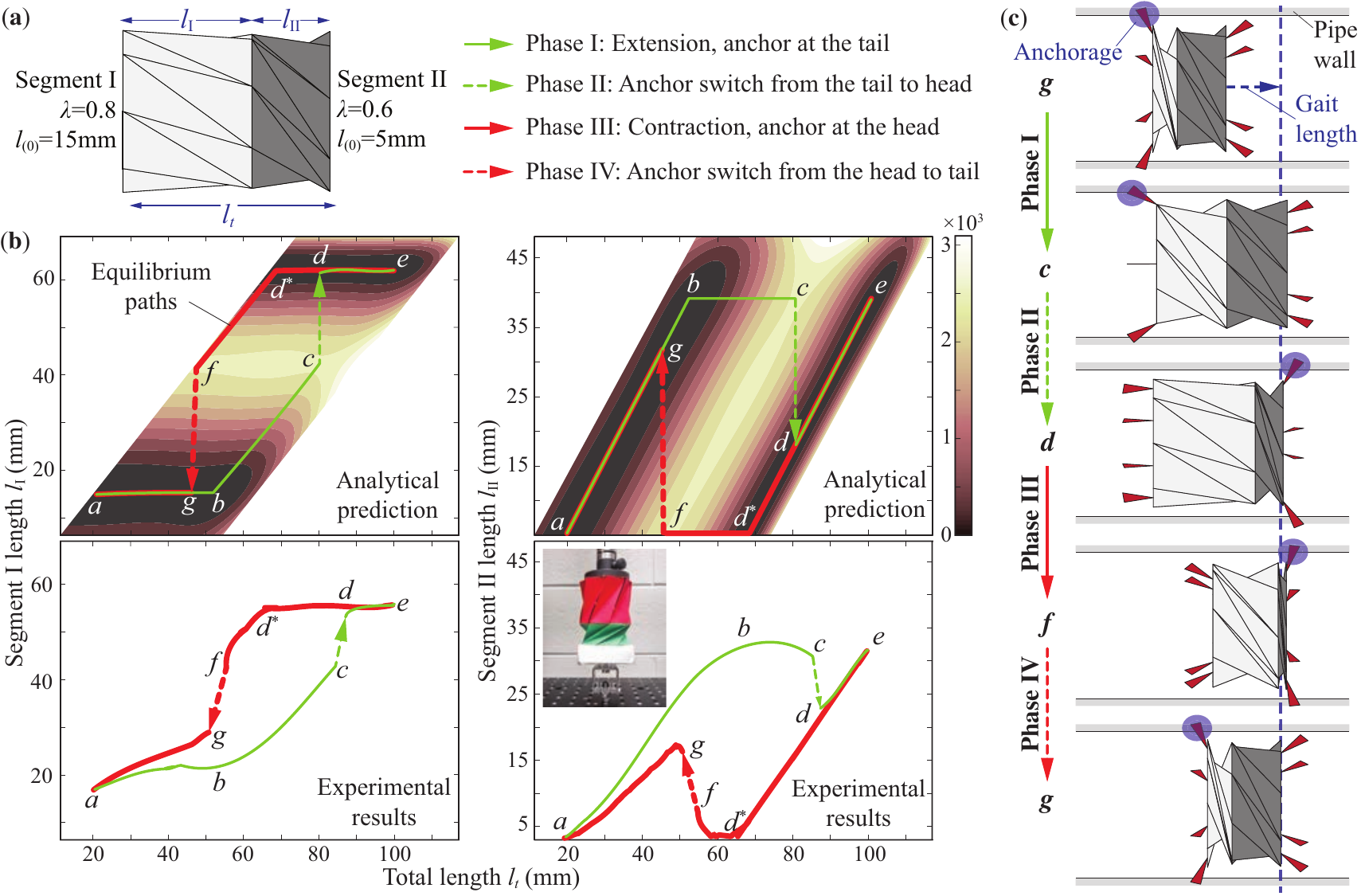}}

\caption{Formation of the actuation cycle in the multi-stable Kresling driving module and the corresponding peristaltic-like locomotion gait.  (a) The design of the driving module and the nomenclature denoting the different phases in the actuation cycle.  (b) Analytical prediction (up) and experimental results (below) of the Segment I and II deformations versus the prescribed change in the total length of the driving module.  The color map represents the total potential energy landscape, where darker color represents lower energy and vice versa.  (d) To-scale schematic diagram of the peristaltic-like crawling gait that is generated using the four phases in the actuation cycle and foldable anchors.  Design parameters of the driving module are listed in Table \ref{table:Module}.}

 \label{fig:ActCycle}
\end{figure}

In this section, we use a case study to illustrate how to harness the multi-stability in Kresling origami to generate a deterministic deformation sequence (or ``actuation cycle'') with only one actuator.  In this case study, the driving module consists of two generalized Kresling segments of different angle ratios and bistability strengths (Figure \ref{fig:ActCycle}(a)).  Without any loss of generality, we assume $\lambda_\text{I}\geqslant\lambda_\text{II}$, where the subscript ``I'' and ``II'' represents the two constituent Kresling segments, respectively.  The Kresling design parameters used for this dual-segment driving module are listed in Table \ref{table:Module}.  To generate the actuation cycle, we stretch and compress this driving module at its two ends without manipulating its two segments individually. That is, we only increase and decrease the {\it total length} ($l_t$) of the driving module without directly controlling the individual segment lengths.

\begin{table}[!ht]
\centering
\caption{Design parameters of the two Kresling segments in the driving module.}
\vspace{5pt}
\begin{tabular}{c c c}
\hline
\textbf{Parameter} & \textbf{Segment I} & \textbf{Segment II}\\
\hline
$N$ & 8 & 8 \\
$P$ (mm) & 30 & 30 \\
$\lambda$ & 0.8 & 0.6 \\
$L_{(0)}$ (mm) & 15 & 5 \\
\hline
\end{tabular}
\label{table:Module}
\end{table}

To identify the actuation cycle, we first need to find how the driving module strain energy changes when the total length ($l_t$) of the driving module is changed from its minimum to maximum and vice versa. This will enable us to get the individual segment deformations and identify the path the system follows as total length ($l_t$) is changed. We applied a customized optimization algorithm to the landscape of total strain energy (Figure \ref{fig:ActCycle}(b)) \cite{Oh2009}. In this optimization, the objective function is the total strain energy $E_{t}=E_\text{I}+E_\text{II}$ according to Equation (\ref{eq:E}). The independent variable is the segment length $l_\text{I}$ or $l_\text{II}$, and they must satisfy the equality constraint $l_\text{I}+l_\text{II}=l_t$, and be within the bounds $l_\text{I min} \leq l_\text{I} \leq l_\text{I max}$, and $l_\text{II min} \leq l_\text{II} \leq l_\text{II max}$.  In this way, the optimization problem becomes: Find the value for $l_\text{I}$ (or $l_\text{II}$) which locally minimizes the scalar objective function $E_t$ for a given a prescribed total length $l_t$, and satisfies the given equality constraint. Results of this optimization are shown as the ``equilibrium paths'' in Figure \ref{fig:ActCycle}(b), and \ref{multi_stab} details a more comprehensive optimization procedure involving multiple Kresling segments.

We start by stretching the driving module when its two segments are both at its fully-contracted stable state (0) (point $a$ in Figure \ref{fig:ActCycle}(b)). During the stretching, the Kresling segments deform by following the equilibrium path $a\rightarrow g\rightarrow  b\rightarrow c\rightarrow d\rightarrow e$ until both of them reach the fully-extended stable state (1) (point $e$).  Then, we compress the driving module and observe that the segments follow a different equilibrium path $e\rightarrow d\rightarrow d^*\rightarrow f\rightarrow g\rightarrow a$ until they come back to the state (0) (Supplemental Video A).

We observe two distinct jumps in these equilibrium paths. One occurs during the stretching from $c \rightarrow  d$, and the other during the compression from $f \rightarrow  g$ (Figure \ref{fig:ActCycle}(b)). When these jumps occur, a branch of local energy minima reaches its end so that the driving module is forced to quickly  ``jump'' to a distant branch of energy minima.  During these jumps, the two Kresling segments change their length significantly, while their total length ($l_t$) remains almost the same.  By combining parts of these equilibrium paths and the two jumps, we can construct an ``actuation cycle'': $g \rightarrow b \rightarrow  c \rightarrow  d \rightarrow d^* \rightarrow  f \rightarrow g$.  This actuation cycle consists of four consecutive ``phases'':  In Phase I ($g \rightarrow b \rightarrow c$), Segment I increases in length significantly while Segment II remains almost fully-extended.  Phase II ($c \rightarrow  d$) is the first jump, by which Segment I quickly reaches the fully-extended state, but Segment II contracts significantly in length.  In Phase III ($d \rightarrow d^* \rightarrow f$), Segment II continues to contract in length until reaching its fully-contracted state, Segment I also contracts but to a lesser degree.   The final Phase IV ($f \rightarrow  g$) is the second jump, by which Segment I quickly deforms to its fully-contracted state, but Segment II extends in length significantly. 

We experimentally verified the formation of this actuation cycle in a paper-based prototype of the driving module (Figure \ref{fig:ActCycle}(b)).  The fabrication procedure and experimental set up are the same as the single Kresling segment tests. The universal testing machine was used to prescribe the change in total length of the driving module ($l_t$). To accurately measure the segment deformation, we obtained high resolution video footage of the driving module and used the MATLAB$^{\circledR}$ Image Processing Toolbox\textsuperscript{TM} to measure the length of Kresling segments ($l_I$ and $l_{II}$). The measured actuation cycle, including the two jumps, agrees well with the analytical predictions.  However, there are some discrepancies between the analytical prediction and experiment measurements.  More specifically, the measured total lengths at which the jumps occur are slightly different from the predictions, and the jumps are not as quick.  Nonetheless, these discrepancies do not hinder the creation of peristaltic-like locomotion as we will discuss in the following section.

\section{Locomotion Gait Generation}
In this section, we show how the actuation cycle, combined with foldable anchors, can create peristaltic-like crawling locomotion.  Segments in the earthworm body increase in diameter while contracting in length and vice versa (Figure \ref{fig:Bigpic}(b)).  This is an important component for achieving peristaltic locomotion because it provides a mechanism to anchor the fully-contracted segment to its surroundings by the setae.  The diameter of Kresling segment, on the other hand, does not change when its length changes. This necessitates the design of anchors which mimic the radial deformation of earthworm segments. 

\subsection{Anchor Design}\label{anchors}
\begin{figure}[h!]
\makebox[\textwidth][c]{\includegraphics{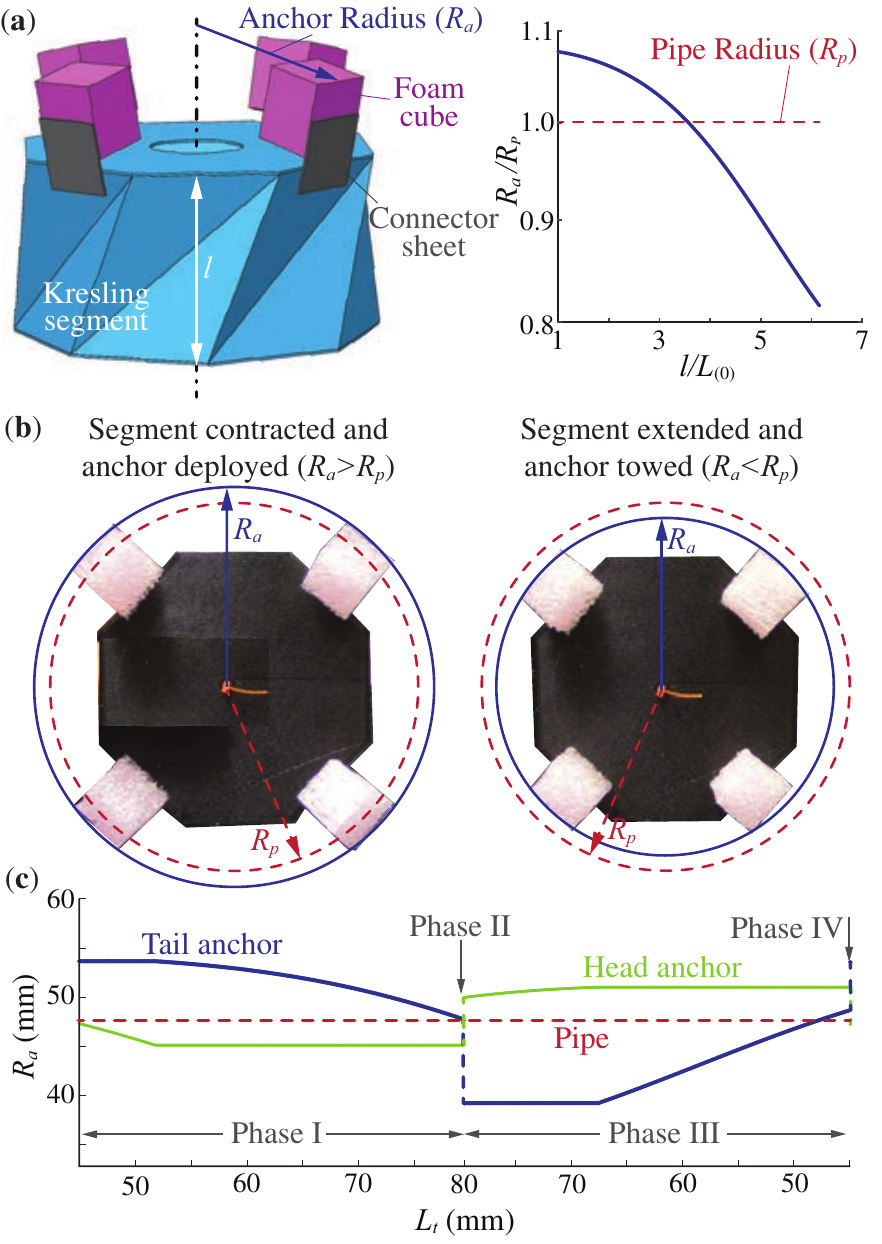}}

\caption{Designs of the foldable anchors:  (a)  Left: a 3D CAD rendering of the Kresling segment with the anchors attached.  Right:  The relationship between the effective anchor radius ($R_a$), pipe radius ($R_p$), and segment length ($l$).  (b) The fabricated segment showing maximum and minimum anchor radius with respect to the pipe radius.  (c) The change in anchor radius during one actuation cycle.  The anchor design parameters are in Table \ref{table:Anchor}.}
 \label{fig:Anchor}
\end{figure}

We designed the anchors by taking advantage of the folding kinematics of Kresling segments.  These anchors are attached to the triangular facets, so they can deploy and increase the effective diameter when the segments are contracting (Figure \ref{fig:Anchor}(a,b)). They have plastic foam cubes at their tips to create sufficient friction and thus a strong anchorage to their surroundings (a pipe of 47.5mm radius in this case). Moreover, we define a ``cut-off'' length for each segment to ensure proper anchor deployment.  When the Kresling segment contracts longitudinally below its cut-off length, its anchors should be deployed far enough to create an anchorage with its surroundings.  For Segment I, its cut-off length is the length at the point $c$ on its equilibrium path as shown in Figure \ref{fig:ActCycle}; for Segment II, its cut-off length corresponds to the point $d$.  We then determined the dimensions of these anchors according to these cut-off lengths, folding kinematics of the Kresling, and the pipe inner diameter (Table \ref{table:Anchor}). The anchors are designated as tail anchor and head anchor according to their position on the robot. The tail anchor is attached to Segment I and the head anchor is attached to the Segment II. In this way, the effective diameter of the Segment I is larger than the pipe diameter during Phase I of the actuation cycle, while the diameter of Segment II is larger than the pipe during Phase III (Figure \ref{fig:Anchor}(c)).  Moreover, the anchoring location switches from Segment I to II in the Phase II jump, and switches back from Segment II to I in the Phase IV jump.  

\begin{table}[]
\centering
\caption{Anchor design parameters, units in mm}
\vspace{5pt}
\begin{tabular}{c c c c}
\hline
\textbf{ } & \textbf{Parameter} & \textbf{Seg. I} & \textbf{Seg. II}\\
\hline
\multirow{3}{4em}{Foam cube} & Length & 18 & 15 \\
&Width  & 15 & 15 \\
&Thickness & 10 & 10 \\
\hline
\multirow{3}{4em}{Connector sheet} & Length & 15 & 15 \\
&Width  & 15 & 15 \\
&Thickness & 0.5 & 0.5 \\
\hline
\end{tabular}

\label{table:Anchor}
\end{table}

\subsection{Peristaltic-like Locomotion Gait}
By combining the dual-segment multi-stable driving module and the properly designed anchors, we now complete the design of crawling robot and harness the actuation cycle to generate a peristaltic-like locomotion gait.  More specifically, the four consecutive phases in the actuation cycle can be used to alternate the anchoring locations between the head and tail of the driving module, resulting in a net forward displacement as detailed below (Figure \ref{fig:ActCycle}(c)):  

In Phase I ($g \rightarrow  b \rightarrow c$), the crawling robot is anchored at its tail because its Segment I is below its cut-off length. Meanwhile, the robot body is increasing in its total length by the actuator input, giving a net forward displacement.

In Phase II ($c \rightarrow d$), the jump between the equilibrium paths switches the anchor location from the tail to the head.  No head or tail displacement occurs during this jump.

In Phase III ($d \rightarrow d^* \rightarrow  f$), the crawling robot is anchored at its head because its Segment II is now below its cut-off length. Meanwhile, the robot body is contracting in its total length, moving the tail forward.

In the final Phase IV ($f \rightarrow g$), the second jump occurs and the anchor location switches back from head to the tail.  There is a slight backward displacement.  At the end of this phase, the crawling robot returns to its original configuration of the actuation cycle, i.e. at the start of Phase I.  The ``gait length'' is the total forward movement of the crawling robot head after one actuation cycle. And the actuation cycle from Phase I to Phase IV can be repeated to drive the robot forward continuously. 

\begin{figure}[h!]
\makebox[\textwidth][c]{\includegraphics{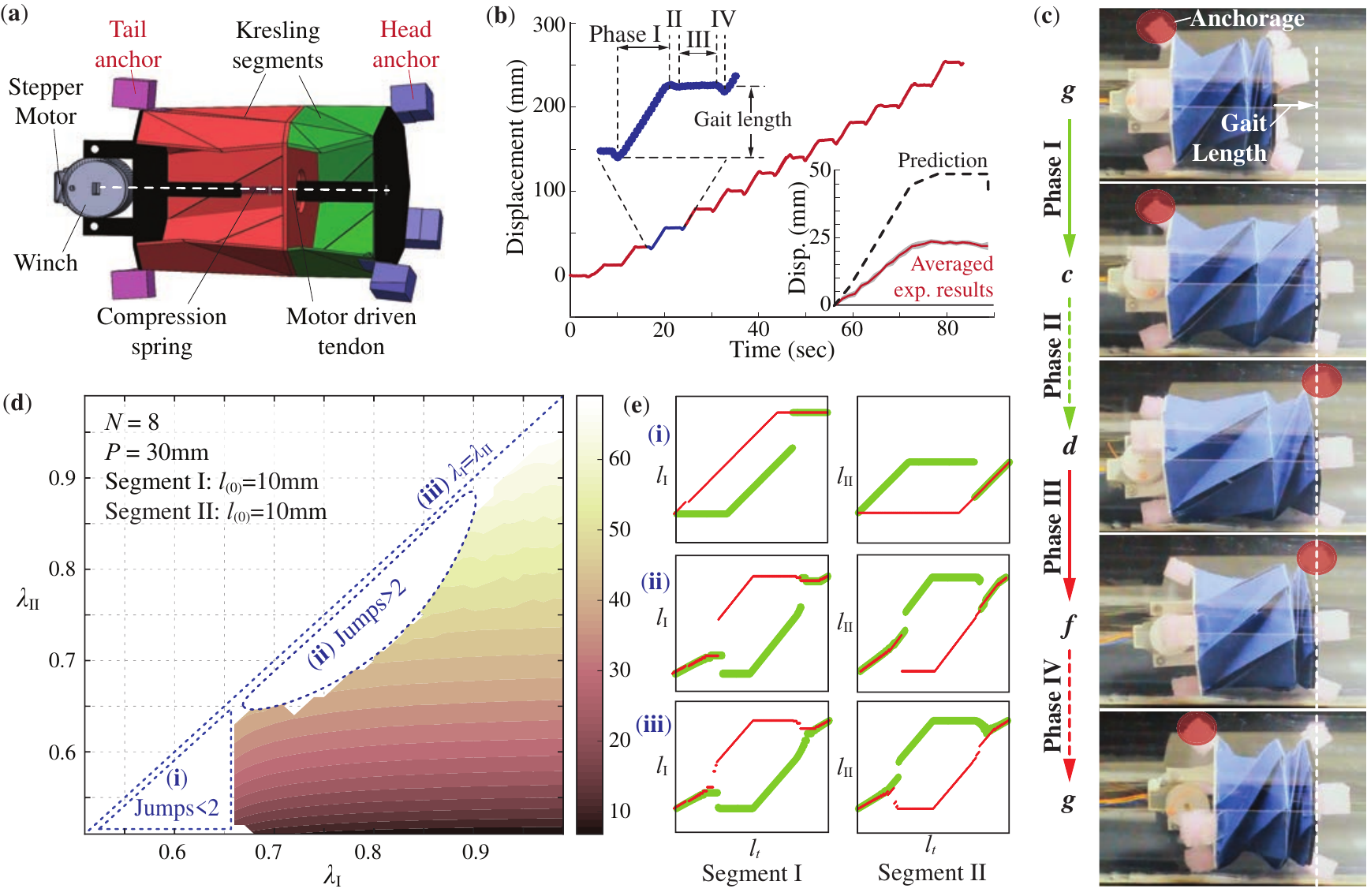}}

\caption{Fabrication and testing of the multi-stable Origami crawler prototype.  (a) 3D CAD rendering of the crawler with a cutaway view showing the motor-winch actuation mechanism.  (b) The measured movement of the robotic head over many actuation cycles.  The insert figure at the upper left corner illustrates the four different phases in one actuation cycle.  Insert at the lower right compares the averaged gait in the experiment and the analytical prediction.  The shaded band is the standard deviation.  (c) The observed four phases of the actuation cycle in the origami crawling robot.  The anchor switch locations and gait length are highlighted.  (d) Results of the parametric study depicting the influence of segment bistability strengths on locomotion gait length.  Here the color map represents the locomotion gait length in millimeters.   (e) Examples of equilibrium paths that do not exhibit any properly defined, four-phased actuation cycles with certain combinations of segment angle ratios.}
 \label{fig:Robot}
\end{figure}

To experimentally validate the peristaltic-like locomotion induced by multi-stability, we fabricated and tested a proof-of-concept prototype of the crawling robot.  This prototype features the same Kresling origami and anchor designs as in the analytical case study (Table \ref{table:Module} and \ref{table:Anchor}).  A compression spring-winch mechanism attached to the two end plates of this robot is used to control its total length (Figure \ref{fig:Robot}(a)).  A 5V stepper motor drives this spring-winch mechanism, and the motor rotation is pre-programmed using Ardruino METRO 328 and motor-shield v2.3. To decrease driving module length, the robot's stepper motor turns the winch, pulling in the attached tendon.  To increase the total length, the motor turns the winch in the opposite direction to release the tendon. The compression spring provides the internal force to keep the tendon taut.  To measure the locomotion performance, we took high-quality video footage of the crawling robot in action and used the Computer Vision Toolbox in MATLAB$^{\circledR}$ (Supplemental Video B).  We developed a computer program using the Kalman filter based motion tracking algorithm to track the movement of the head of the robot.

The experimental results summarized in Figure \ref{fig:Robot}(b,c) agree quite well with the analytical predictions in Figure \ref{fig:ActCycle}(c) regarding the segment deformation sequence and anchor location switches. Moreover, the robot locomotion cycle is uniform and repetitive (Figure \ref{fig:Robot}(b)).  There is a discrepancy regarding the magnitude of gait length between the experiment and analysis, and two factors can contribute to this.  One is that the analytical prediction uses an idealistic model to characterize the Kresling bi-stability so it does not fully capture the behaviors of the physical prototypes (also evident in Figure \ref{fig:ActCycle}(b)).  The other factor is the slippage between the pipe and robot anchors, which results from the temporary loss of contact during the anchor switching in Phase II and IV.  Regardless, this experiment firmly validates the feasibility of using multi-stability in Kresling origami to create the peristaltic-like locomotion with only one actuator and without a complex control architecture to coordinate the segments. That is, the deformation of the segments and anchorage locations are ``coordinated'' directly by the mechanics of multi-stability.  
 
\begin{table}[]
\centering
\caption{Features and performance of the final origami crawler prototype}
\vspace{5pt}
\begin{tabular}{l c}
\hline
\textbf{Feature} & \textbf{Value}\\
\hline
Mass & 70 g\\
Maximum length & 90 mm \\
Minimum length & 55 mm \\
Average speed  & 3.3 mm/sec \\
Average gait length & 22 mm \\
Average cycle duration & 6.7 sec\\
\hline
\end{tabular}
\label{table:Feature}
\end{table}

Table \ref{table:Feature} summarizes the features and locomotion performances of the dual-segment multi-stable origami crawler.  It is important to highlight that the actuation cycle induced by multi-stability is independent of the {\it rate} of stretching/compression in total length.  Therefore, by changing the rotational speed of the motor one can adjust the {\it frequency} of the locomotion cycle and thus the averaged crawling speed, however, the motor speed does not affect the gait length in one locomotion cycle.  The gait length is only related to the Kresling origami design and the corresponding multi-stability. We detail this further in the following parametric study.

\subsection{Parametric Study: Gait Length}\label{parastudy}
To uncover the correlations between peristalsis gait length and multi-stability, we performed a parametric study by combining two Kresling segments of different bistability strength (aka. different angle ratios $\lambda$).  Without any loss of generality, let's assume that $\lambda_\text{I}\geqslant\lambda_\text{II}$.  It is clear from the actuation cycle study that the locomotion gait length depends both on the magnitude of jumps and the segment deformations between the two jumps (Figure \ref{fig:ActCycle}(b)).  Results of the parametric study show that, for a given $\lambda_\text{I}$, gait length increases as $\lambda_\text{II}$ increases. On the other hand, for a given $\lambda_\text{II}$, gait length is insensitive to changes in $\lambda_\text{I}$ (Figure \ref{fig:Robot}(d)).   There are three possible scenarios by which peristaltic-like locomotion is unachievable.  First, when both $\lambda_\text{I}$ and $\lambda_\text{II}$ are smaller than 0.65, their bistability strengths are too weak to induce any jumps or anchor location switches (case (i) in Figure \ref{fig:Robot}(e)).  Second, when the two $\lambda$ values are very close to each other, we observe multiple jumps in the actuation cycle so peristaltic locomotion is not possible (case (ii) in Figure \ref{fig:Robot}(e)).  Lastly, when $\lambda_\text{I}=\lambda_\text{II}$, there are no discernible jumps that can create any actuation cycles (case (iii) in Figure \ref{fig:Robot}(e)).

\section{Summary and Conclusion}

In this study, we demonstrated the use of multi-stability embedded in a robotic origami skeleton to create peristaltic-like locomotion without the need for multiple actuators or complicated controllers.   By combining two bistable Kresling origami segments into a driving module and increasing/decreasing its total length, one can generate a deterministic deformation sequence (or actuation cycle).  This actuation cycle has two discrete ``jumps'' that can significantly change the length of two constituent Kresling segments without affecting their total length.  These jumps are the result of the folding-induced multi-stability, and they naturally divide the actuation cycle into four distinct phases.  We then designed and experimentally validated a peristaltic-like robotic crawling by using two phases for moving the robot forward and the other two for switching the anchoring locations.  To ensure proper anchorage to the surroundings, we designed and implemented foldable anchors according to the kinematics of Kresling folding.  Results of this work show that the mechanics of multi-stability can be used to directly coordinate the robotic motion and drastically simplify the mechatronic setup and control of compliant robots.  Essentially, the mechanics of multi-stability can impart ``intelligence'' to the robotic body so that it can take up the low-level control task of locomotion generation.  

While we have used a compression spring-winch based linear actuator to control the length of the driving module, any other mechanism that can work in the required deformation range may be used to actuate the robot. The scale independence of the origami mechanism ensures that the same robot design principles can be used to create nano/micro-scale as well as large-scale robots. Moreover, it is worth highlighting that although Kresling origami is used in this study for its simplicity and versatility, the principle of using multi-stability to generate peristaltic-like locomotion is applicable to any other segmented robot systems, as long as the segment can (i) exhibit a coupled longitudinal and radial deformations (aka. expanding radially while shrinking longitudinally, and vice versa) and (ii) exhibit a tunable bistability. Therefore, results of this study can foster a new family of robots that use the mechanics of multi-stability for crawling locomotion.

\section{Acknowledgements}
The authors acknowledge the support from the National Science Foundation (Award \# CMMI-1633952, 1751449 CAREER) and Clemson University (via startup funding and the CECAS Dean’s Faculty Fellow Award).


\appendix

\section{Equilibrium Paths Search}\label{multi_stab}
In this section, we elucidate how to calculate the equilibrium paths of a driving module consisting of any number of serially connected bistable segments.  We calculate the ``equilibrium path'' followed by the driving module via searching for its {\it local potential energy minima} at a prescribed total length. The total potential energy ($E_{t}$) and total length ($l_{t}$) of a driving module of $n$ bistable segments are defined as,

\begin{equation}
E_{t}=\sum_{i=1}^n E_i(l_i)\textrm{, and } l_{t}=\sum_{i=1}^n l_i,
\end{equation}

\noindent where $l_i$ is current length of the $i^\text{th}$ segment and $E_i$ is the corresponding strain potential energy.

The search for equilibrium paths can be defined as an optimization problem.  The objective function of this optimization is the total strain energy $E_{t}$ (a scalar function), and it is to be minimized over the $\mathcal{R}^{n-1}$ vector space of individual segment lengths $\mathbf{L}=[l_1 \dots l_{n-1}]$. An equality constraint regarding the prescribed total length must be satisfied so that $l_n=l_{t}-\sum_{i=1}^{n-1} l_i$. Therefore, the optimization problem can be described as follows: 
\begin{align}
& \textrm{Minimize  }
E_{t}=\sum_{i=1}^{n-1}E_i(l_i) + E_n(l_{t}-\sum_{i=1}^{n-1} l_i)  \nonumber\\
& \textrm{satisfying the bounds  } l_{i \text{ min}} \leq l_i \leq l_{i \text{ max}} \textrm{  and  } l_{n \text{ min}} \leq l_n \leq l_{n \text{ max}}, \nonumber\\
&  \textrm{where  } i=1,2 ... n-1
\label{eq:optim}
\end{align}
The optimization problem described in equation \ref{eq:optim} is solved for every prescribed total length of the driving module $l_{t}$, which is increased from its minimum to the maximum value with an incremental step $\Delta l_{t}$,
\begin{align}
&  l_{t}^\text{min}=\sum_{i=1}^{n} l_{i \text{ (0)}}, \nonumber \\
&  l_{t}^\text{max}=\sum_{i=1}^{n} l_{i \text{ (1)}}, \nonumber \\
&\textrm{and  }m=\frac{(l_{t}^\text{max}-l_{t}^\text{min})}{\Delta l_{t}},
\end{align}
where $m$ is the total number of increments. $l_{i \text{ (0)}}$ and $l_{i \text{ (1)}}$ are the $i^\text{th}$ segment's length at its fully-contracted stable state (0) and fully-extended stable state (1), respectively. Notice that $l_{i \text{ (0)}}$ is different from $l_{i \text{ min}}$ by definition, and typically $l_{i \text{ min}}<l_{i \text{ (0)}}$.  Similarly $l_{i \text{ max}}>l_{i \text{ (1)}}$.  

For the $j^{\textrm{th}}$ increment in $l_t^j$ ($j =2 ... m$), the solutions of the optimization problem are the vectors of individual segment lengths $\mathbf{L}^j=[l_i^j ... l_{n-1}^j]$ corresponding to a local minima of $E_t$. The following pseudo-code describes the optimization algorithm used to search for the equilibrium path of driving module {\it when it is stretched from} $l_t^\text{min}$.

{\bf Step 1}: Initialize the optimization problem using the segment lengths at their fully-contracted stable states (aka. $l_{i \text{ (0)}}$). Thus, we do not have to perform optimization for this first increment because it already corresponds to an energy minima. For segments $i=1,2 ... n-1$ define,
\begin{align}
&     l_i^1=l_{i \text{ (0)}} \textrm{ and  } l_{t}^1=l_{t}^\text{min} \nonumber\\
& \mathbf{L}^1=[l_1^1 \dots l_{n-1}^1].
\end{align}
{\bf Step 2}: Initiate the next increment step $j=2$ so that,
\begin{align}
&      l_{t}^2=l_{t}^1+\Delta l_{t} \nonumber\\
& \mathbf{L}_0^2=[l_1^1 \dots   l_{n-1}^1]=\mathbf{L}^1.
\end{align}
Here, $l_{t}^2$ is the total length of driving module at this increment step, and $\mathbf{L}_{0}^2$ is the initial input needed to solve the optimization problem. 

{\bf Step 3}: Solve the optimization problem described in Equation \ref{eq:optim}. Here, $\mathbf{L}_{0}^2$ is the first guess for finding the optimized lengths of individual segments and the optimized result is recorded as $\mathbf{L}^2$. The length of the last segment is calculated as,
\begin{align}
& l_n^2=l_{t}^2-\sum_{i=1}^{n-1}l_i^2.
\label{eq:sol1}
\end{align}

The corresponding total energy of the module is then written as,
\begin{align}
E_{t}^\text{2 min}=\sum_{i=1}^{n}E_i^2(l_i^2)
\label{eq:sol2}
\end{align}

{\bf Step 4}: Prepare the next increment step by setting $\mathbf{L}_0^{j+1}=\mathbf{L}^j$ and $l_t^j=l_t^\text{min}+(j-1)\Delta l_t$, ($j=3...m$). Then solve the optimization problem again using the procedures in Step 3.  Notice that the optimization output $\mathbf{L}^j$ in the $j^\text{th}$ increment step is always used as the initial guess $\mathbf{L}_0^{j+1}$ of the next $(j+1)^\text{th}$ increment step. 

{\bf Step 5}: Repeat Steps 2 to 4 until $j=m$. 

Equations \ref{eq:sol1} and \ref{eq:sol2} together provide the segment lengths and potential energy along the equilibrium path of this driving module when it is stretched from the minimun length $l_t^\text{min}$. The equilibrium path for compressing the driving module need not be same as the path for stretching it. Thus, the similar procedure can be used to search for the other equilibrium path of the driving module when it is compressed from the maximum length $l_t^\text{max}$.

\end{document}